# GENERATING PRECONDITION EXPRESSIONS IN INSTRUCTIONAL TEXT

**Keith Vander Linden**
ITRI, University of Brighton
Lewes Road
Brighton, BN2 4AT
UK
Internet: knvl@itri.bton.ac.uk

## Abstract

This study employs a knowledge intensive corpus analysis to identify the elements of the communicative context which can be used to determine the appropriate lexical and grammatical form of instructional texts. IMAGENE, an instructional text generation system based on this analysis, is presented, particularly with reference to its expression of precondition relations.

## INTRODUCTION

Technical writers routinely employ a range of forms of expression for precondition expressions in instructional text. These forms are not randomly chosen from a pool of forms that say "basically the same thing" but are rather systematically used based on elements of the communicative context. Consider the following expressions of various kinds of procedural conditions taken from a corpus of instructional text:

(**1a**) *If light flashes red,* insert credit card again. (Airfone, 1991) [1]

(**1b**) *When the 7010 is installed* and *the battery has charged for twelve hours*, move the OFF/STBY/TALK [8] switch to STBY. (Code-a-phone, 1989)

(**1c**) The BATTERY LOW INDICATOR will light *when the battery in the handset is low.* (Excursion, 1989)

(**1d**) Return the OFF/STBY/TALK switch to STBY *after your call.* (Code-a-phone, 1989)

(**1e**) *1. Make sure the handset and base antennas are fully extended.* 2. Set the OFF/STBY/TALK SWITCH to Talk. (Excursion, 1989)

As can be seen here, procedural conditions may be expressed using a number of alternative lexical and grammatical forms. They may occur either before or after the expression of their related action (referred to here as the issue of *slot*), and may be linked with a variety of conjunctions or prepositions (the issue of *linker*). Further, they may be expressed in a number of grammatical forms, either as actions or as the relevant state brought about by such actions (called here the *terminating* condition). Finally, they may or may not be combined into a single sentence with the expression of their related action (the issue of *clause combining*).

Text generation systems must not only be capable of producing these forms but must also know when to produce them. The study described here has employed a detailed corpus analysis to address these issues of choice and has implemented the results of this study in IMAGENE, an architecture for instructional text generation.

## CORPUS ANALYSIS

The corpus developed for this study contains approximately 1000 clauses (6000 words) of instructions taken from 17 different sources, including instruction booklets, recipes, and auto-repair manuals. It contains 98 precondition expressions, where the notion of precondition has been taken from Rhetorical Structure Theory (Mann and Thompson, 1988), and in particular from Rösner and Stede's modified relation called *Precondition* (1992). This relation is a simple amalgam of the standard RST relations Circumstance and Condition and has proven useful in analyzing various kinds of conditions and circumstances that frequently arise in instructions.

The analysis involves addressing two related issues:

1. Determining the range of expressional forms commonly used by instructional text writers;

2. Determining the precise communicative context in which each of these forms is used.

---

[1] In this paper, a reference will be added to the end of all examples that have come directly from the corpus, indicating the manual from which they were taken.

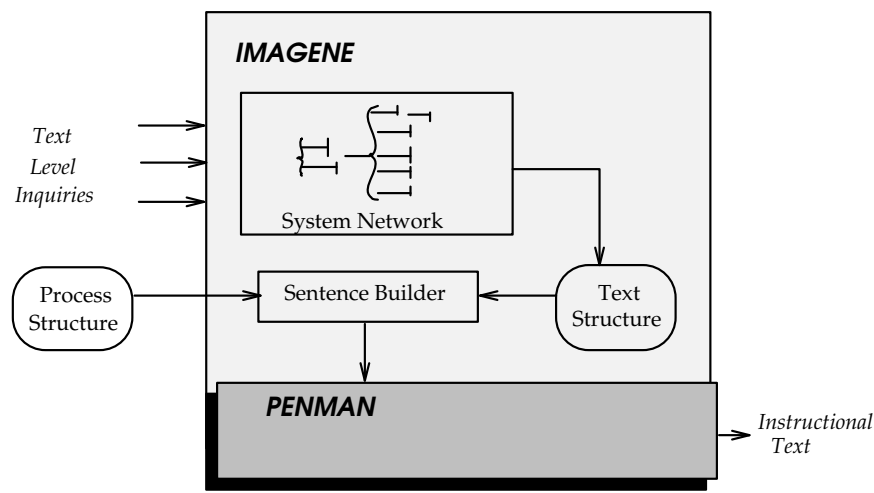

Figure 1: The Architecture of IMAGENE

Determining the range of forms was a matter of cataloging the forms that occurred in the corpus. Example (1) shows exemplars of the major forms found, which include present tense action expressions (1a), agentless passives (1b), relational expressions of resultant states (1c), phrasal forms, (1d), and separated imperative forms (1e).

Determining the functional context in which each of the forms is used involves identifying correlations between the contextual features of communicative context on the one hand, and the lexical and grammatical form on the other. I focus here on the range of lexical and grammatical forms corresponding to the precondition expressions in the corpus. The analyst begins by identifying a feature of the communicative context that appears to correlate with the variation of some aspect of the lexical and grammatical forms. They then attempt to validate the hypothesis by referring to the examples in the corpus. These two phases are repeated until a good match is achieved or until a relevant hypothesis cannot be found.

## IMAGENE

The analysis has resulted in a number of identified covariations which have been coded in the System Network formalism from Systemic-Functional Linguistics (Halliday, 1976) and included in the IMAGENE architecture. The system network is basically a decision network where each choice point distinguishes between alternate features of the communicative context. It has been used extensively in Systemic Linguistics to address both sentence-level and text-level issues. Such networks are traversed based on the appropriate features of the communicative context, and as a side-effect of this traversal, linguistic structures are constructed by *realization statements* which are associated with each feature of the network. These statements allow several types of manipulation of the evolving text structure, including the insertion of text structure nodes, grammatical marking of the nodes, textual ordering, and clause combining. Currently, the network is traversed manually; the data structures and code necessary to automatically navigate the structure have not been implemented. This has allowed me to focus on the contextual distinctions that need to be made and on their lexical and grammatical consequences.

The general architecture of IMAGENE, as depicted in Figure 1, consists of a System Network and a Sentence Building routine, and is built on top of the Penman text generation system (Mann, 1985). It transforms inputs (shown on the left) into instructional text (shown on the right).

The following sections will detail the results of the analysis for precondition expressions. It should be noted that they will include intuitive motivations for the distinctions made in the system network. This is entirely motivational; the determinations made by the systems are based solely on the results of the corpus analysis.

## PRECONDITION SLOT

In the corpus, preconditions are typically fronted, and therefore the sub-network devoted to precondition expression will default to fronting. There are four exceptions to this default which are illustrated here:

(2a) The BATTERY LOW INDICATOR will light *when the battery is the handset is low.* (Excursion, 1989)

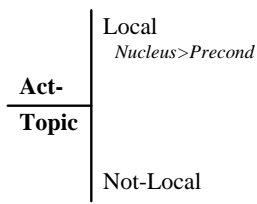

Figure 2: The Precondition Slot Selection Network

**(2b)** Return the OFF/STBY/TALK switch to STBY *after your call.* (Code-a-phone, 1989)

**(2c)** The phone will ring *only if the handset is on the base.* (Code-a-phone, 1989)

**(2d)** In the STBY (standby) position, the phone will ring *whether the handset is on the base or in another location.* (Code-a-phone, 1989)

The slot selection for example (2a) could go either way, except that it is the first sentence in a section titled "Battery Low Indicator", making the discussion of this indicator the local topic of conversation, and thus the appropriate theme of the sentence. This distinction is made in the portion of the system network shown in figure 2. This sub-network has a single system which distinguishes between preconditions associated with actions referring to thematic material and those associated with non-thematic material. The realization statement, *Nucleus>Precond*, indicates that the main action associated with the condition (called the nucleus in RST terminology) is to be placed before the precondition itself.

The slot determinations for the remainder of example (2) are embedded in system networks shown later in this paper. Example (2b) is the example of what I call *rhetorical demotion*. The action is considered obvious and is thus demoted to phrase status and put at the end of its immediately following action. Examples (2c) and (2d) show preconditions that are not fronted because of the syntax used to express the logical nature of the precondition. In (2c), the condition is expressed as an exclusive condition which is never fronted. One could, perhaps, say "?? Only if the handset is on the base, will the phone ring.",[2] but this form is never used in the corpus. Neither is the condition form in (2d) ever fronted in the corpus.

---

[2] The "??" notation is used to denote a possible form of expression that is not typically found in the corpus; it does not indicate ungrammaticality.

## PRECONDITION LINKER

Preconditions are marked with a number of linkers, illustrated in the following examples:

**(3a)** Lift the handset and set the OFF/STBY/TALK [8] switch to TALK. *When* you hear dial tone, dial the number on the Dialpad [4]. (Code-a-phone, 1989)

**(3b)** *If* you have touch-tone service, move the TONE/PULSE SWITCH to the Tone position. (Excursion, 1989)

**(3c)** *1. Make sure* the handset and base antennas are fully extended. 2. Set the OFF/STBY/TALK SWITCH to Talk. (Excursion, 1989)

The systems largely dedicated to selecting precondition linkers are shown in figure 3.[3] Two parallel systems are entered, **Condition-Probability** and **Changeable-Type**.

**Condition-Probability** distinguishes actions which are probable from those which are not. Highly probable actions are typically marked with "when". Those actions which are not highly probably are marked with "If" or some similar linker, as determined by the **Complexity** system and its descendants.

The **Complexity** system is entered for actions which are not probable and not changeable. It determines the logical nature of the preconditions and sets the linker accordingly. The three possible linkers chosen by this sub-network are "if", "only if", or "whether ... or ... ".

**Precond-When** is entered when the action is conditional and further is highly probable. The occurrence of the dial tone in example (3a) is part of a sequence of actions and is conditional in that it may not actually happen, say if the telephone system is malfunctioning in some way, but is nonetheless highly probable. **Precond-Nominal** is entered immediately after **Precond-When** whenever the precondition is being stated as a nominalization. It overwrites the linker choice with "after" in only this case.

Preconditions that the user is expected to be able to change if necessary and which come at the beginning of sections that contain sequences of prescribed actions are called *Changeable* preconditions. Example (3c) is such a case. Here, the reader is expected to check the antennas and extend them if they are not already extended. This type of precondition is marked as a "Make sure" imperative clause by **Changeable-Type**.

---

[3] In the figure, the bold-italic conditions attached to the front of these systems denote conditions that hold on entry (e.g., Conditional-Action is a condition true on the entry of Condition-Probability). They are necessary because the networks shown are only portions of a much larger network.

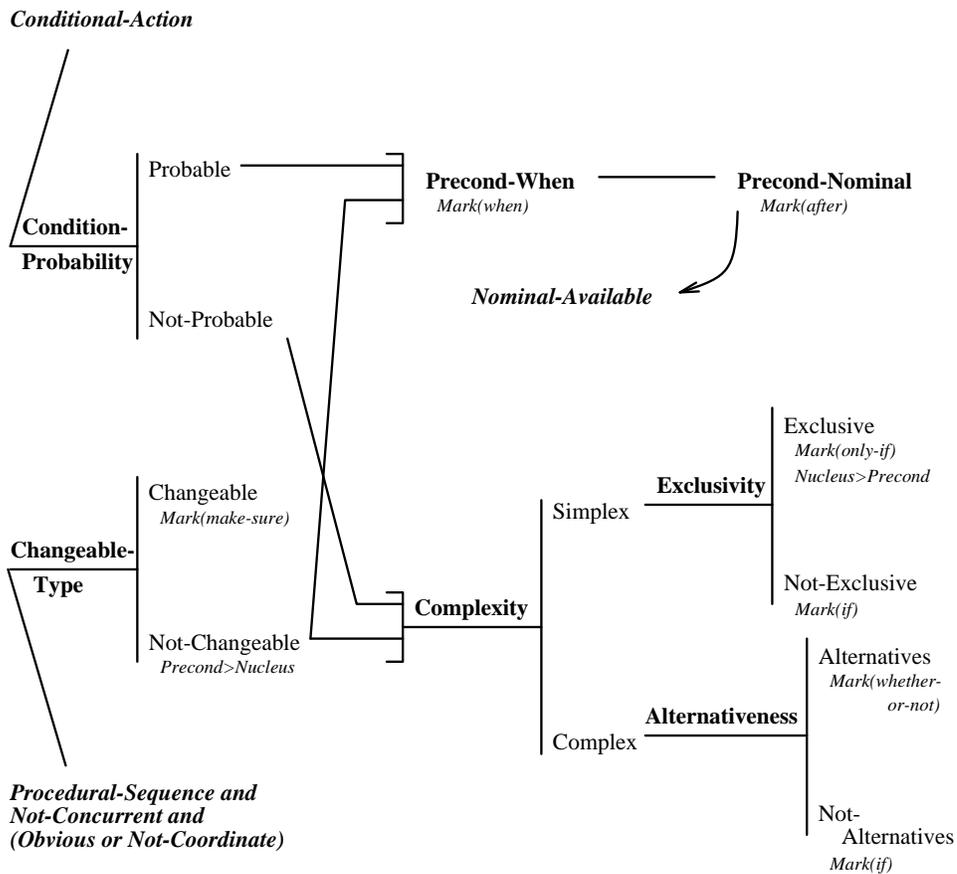

Figure 3: The Precondition Linker Selection Network

## PRECONDITION FORM

As noted above, preconditions can be expressed as either a terminating condition or as an action. The choice between the two is made by the form selection sub-networks, shown in figures 4 and 5. This choice depends largely upon the type of action on which the precondition is based. The actions in the corpus can be divided into five categories which affect the grammatical form of precondition expressions:

- Monitor Actions;
- Giving Actions;
- Placing Actions;
- Habitual Decisions;
- Other Actions.

The first four actions are special categories of actions that have varying act and terminating condition forms of expression. The last category, other actions, encompasses all actions not falling into the previous four categories. The sub-network which distinguishes these forms is shown in figure 4. This section will discuss each category in turn, starting with the following examples of the first four action types:

(**4a**) *Listen for dial tone,* then dial AREA CODE + NUMBER slowly. (Airfone, 1991)

(**4b**) *If you have touch-tone service*, move the TONE/PULSE SWITCH to the Tone position. (Excursion, 1989)

(**4c**) The phone will ring *only if the handset is on the base.* (Code-a-phone, 1989)

(**4d**) *If you leave the OFF/STBY/TALK [8] switch in TALK*, move the switch to PULSE, and tap FLASH [6] the next time you lift the handset, to return to PULSE dialing mode. (Code-a-phone, 1989)

Monitor actions, as shown in example (4a), concern explicit commands to monitor conditions in the environment. In this case, readers are being commanded to listen for a dial tone, with the underlying assumption that they will not continue on with the instructions unless one is heard. Giving and Placing actions, however, tend to be expressed as terminating conditions, as shown in (4b) and (4c). The corpus does not include active forms of these actions, such as "?? *If the phone company has given you touch-tone service,* do ..." or "?? Do ... *if you have placed the handset on the base.*" An *Habitual decision* is a decision to make a practice of performing some action or of performing an action in some way. When stated as preconditions, they take the present tense form in (4d). Taken in context, this expression refers not to a singular action of leaving the OFF/STBY/TALK switch in TALK position, but rather to the decision to habitually leave it in such a state. The singular event would be expressed as "*If you have left the OFF/STBY/TALK switch in TALK,* do ..." which means something quite different from the expression in (4d) which is stated in present tense.

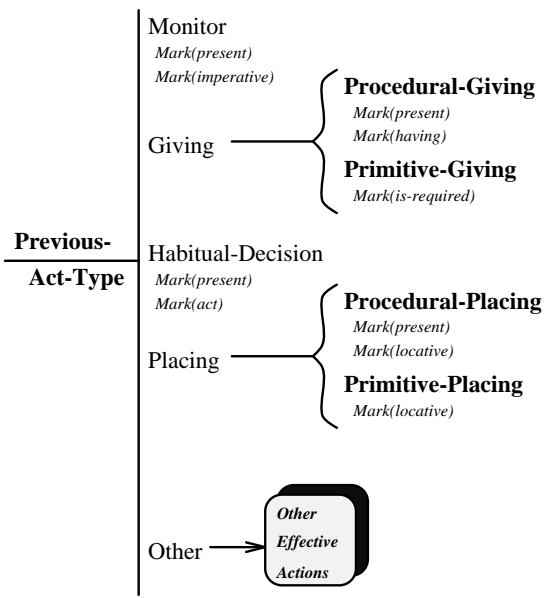

Figure 4: The Precondition Form Selection Network

The bulk of the preconditions in the corpus (70.4%) are based on other types of actions. These types are distinguished in figure 5. In general, the Other Effective Action systems are based on the actor of the action. Reader actions are expressed either as present tense passives or as present tense actions, depending upon whether the action has been mentioned before or not. These distinctions are made by the gates **Repeated-Reader** and **Not-Repeated-Reader**. An example of the former can be found in (5a), ("When the 7010 is installed"). In the corpus, such expressions of actions already detailed in the previous text take the present tense, agentless passive form. If the reader action is not a repeated mention, a simple present tense active form is used, as in example (5b).

(**5a**) *When the 7010 is installed* and the battery has charged for twelve hours, move the OFF/STBY/TALK [8] switch to STBY. (Code-a-phone, 1989)

(**5b**) *If you make a dialing error*, or want to make another call immediately, FLASH gives you new dial tone without moving the OFF/STBY/TALK switch. (Code-a-phone, 1989)

The **Act-Hide** system and its descendants are entered for non-obvious, non-reader actions. There are four basic forms for these precondition expressions, examples of which are shown here:

(**6a**) *If light flashes red,* insert credit card again (Airfone, 1991)

(**6b**) *When you hear dial tone*, dial the number on the Dialpad [4]. (Code-a-phone, 1989)

(**6c**) The BATTERY LOW INDICATOR will light *when the battery in the handset is low.* (Excursion, 1989)

(**6d**) *When instructed (approx. 10 sec.)* remove phone by firmly grasping top of handset and pulling out. (Airfone, 1991)

**Act-Hide** distinguishes actions which are overly complex or long duration and those that are not. Those which are not will be expressed either as present tense actions, as the one in example (6a), if the action form is available in the lexico-grammar. **Active-Available** makes this determination. If no action form is available, then **Inception-Status** is entered. If the inception of the action is expected to have been witnessed by the reader, then the present tense sensing action form is used, as shown in example (6b).

**Termination-Availability** is entered either if the action is to be hidden or if the inception of the action was not expected to be experienced by the reader. In these cases, the relational form of the terminating condition is used if it is available. An example of this is shown in example (6c). The long duration action of the battery draining is not expressed in the relational form used there. If the relational form is not available, the present tense, agentless passive is specified, as shown in example (6d).

Finally, if an action being expressed as a precondition is considered obvious to the reader, the nominalization is used, provided its nominalized form is available in the lexicon. Example (1d) is an example of such an expression.

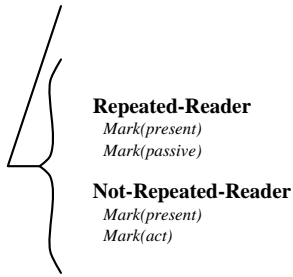
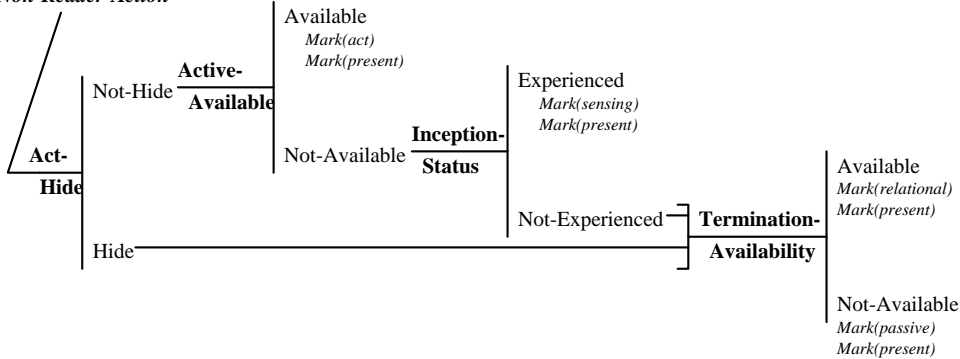

Figure 5: The Other Effective Actions Selection Network

## VERIFYING IMAGENE'S PRESCRIPTIONS

This study has been based primarily on an analysis of a small subset of the full corpus, namely on the instructions for a set of three cordless telephone manuals. This training set constitutes approximately 35% of the 1000 clause corpus. The results of this analysis were implemented in IMAGENE and tested by manually re-running the system network for all of the precondition expressions in the training set. These tests were performed without the Penman realization component engaged, comparing the text structure output by the system network with the structure inherent in the corpus text. A sample of such a text structure, showing IMAGENE's output when run on the actions expressed in the text in example (7), is shown in figure 6. The general structure of this figure is reflective of the underlying RST structure of the text. The nodes of the structure are further marked with all the lexical and grammatical information relevant to the issues addressed here.

(7) *When the 7010 is installed* and the battery has charged for twelve hours, move the OFF/STBY/TALK [8] switch to STBY. The 7010 is now ready to use. Fully extend the base antenna [12]. Extend the handset antenna [1] for telephone conversations. (Code-a-phone, 1989)

Statistics were kept on how well IMAGENE's text structure output matched the expressions in the corpus with respect to the four lexical and grammatical issues considered here (i.e., slot, form, linker, and clause combining). In the example structure, all of the action expressions are specified correctly except for the **Charge** action (the second clause). This action is marked as a present tense passive, and occurs in the corpus in present perfect form.

In full realization mode, IMAGENE translates the text structure into sentence generation commands for the Penman generation system, producing the following output for example (7):

(8) *When the phone is installed, and the battery is charged, move the OFF/STBY/TALK switch to the STBY position. The phone is now ready to use. Extend the base antenna. Extend the handset antenna for phone conversation.*

As just mentioned, this text identical to the original with respect to the four lexical and grammatical issues addressed in the corpus study with the exception of the second clause. There are other differences, however, having to do with issues not addressed in the study, such as referring expressions and the expression of manner. A corpus study of these issues is yet to be performed.

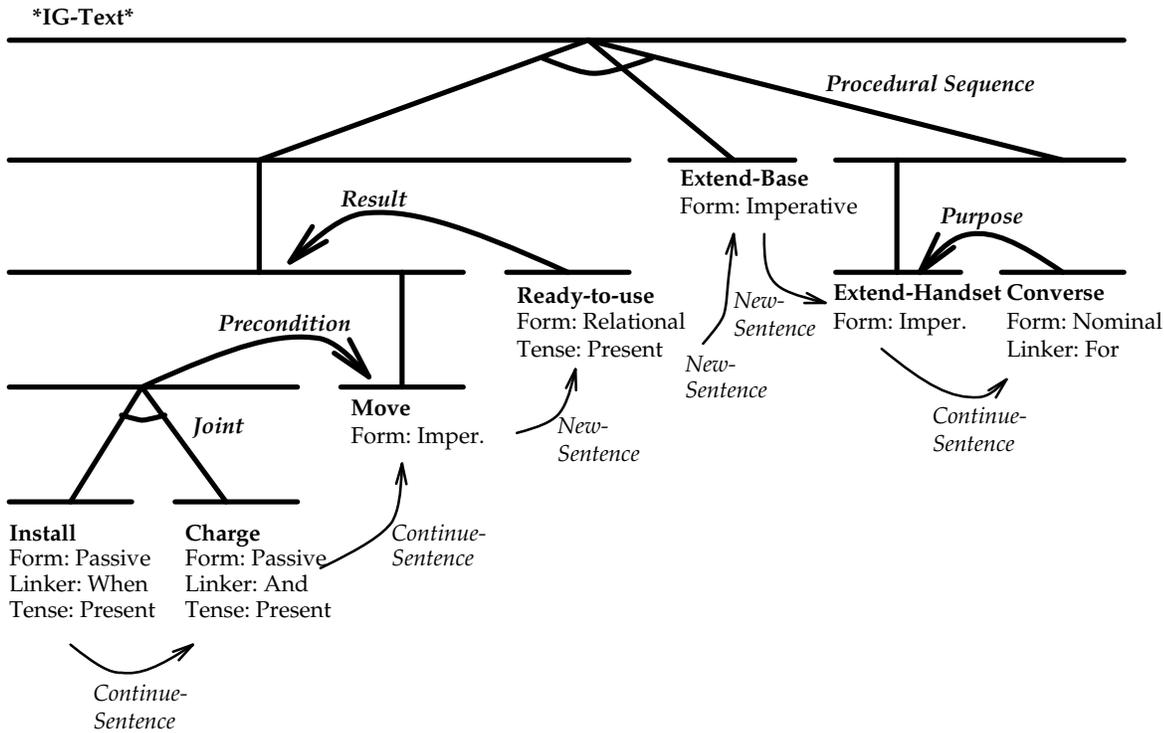

Figure 6: A Sample Text Structure

The overall results are shown in table 7 (see Vander Linden, 1993b for the results concerning other rhetorical relations). This chart indicates the percentage of the precondition examples for which IMAGENE's predictions matched the corpus for each of the four lexical and grammatical issues considered. The values for the training and testing sets are differentiated. The training set results indicate that there are patterns of expression in cordless telephone manuals that can be identified and implemented.

The system's predictions were also tested on a separate and more diverse portion of the corpus which includes instructions for different types of devices and processes. This additional testing serves both to disallow over-fitting of the data in the training portion, and to give a measure of how far beyond the telephone domain the predictions can legitimately be applied. As can be seen in figure 7, the testing set results were not as good as those for the training set, but were still well above random guesses.

Figure 7: The Accuracy of IMAGENE's Realizations for Precondition Expressions — REMOVED

## CONCLUSIONS

This study has employed a knowledge intensive corpus analysis to identify the elements of the communicative context which can be used to determine the appropriate lexical and grammatical form of precondition expressions in instructional texts. The methodology provides a principled means for cataloging the use of lexical and grammatical forms in particular registers, and is thus critical for any text generation project. The current study of precondition expressions in instructions can be seen as providing the sort of register specific data required for some current approaches to register-based text generation (Bateman and Paris, 1991).

The methodology is designed to identify co-variation between elements of the communicative context on the one hand and grammatical form on the other. Such covariations, however, do not constitute proof that the technical writer actually considers those elements during the generation process, nor that the prescribed form is actually more effective than any other. Proof of either of these issues would require psycholinguistic testing. This work provides detailed prescriptions concerning how such testing could be performed, i.e., what forms should be tested and what contexts controlled for, but does not actually perform them (cf. Vander Linden, 1993a).

The analysis was carried out by hand (with the help of a relational database), and as such was tedious and limited in size. The prospect of automation, however, is not a promising one at this point. While it might be possible to automatically parse the grammatical and lexical forms, it remains unclear how to automate the determination of the complex semantic and pragmatic features relevant to choice in generation. It might be possible to use automated learning procedures (Quinlan, 1986) to construct the system networks, but this assumes that one is given the set of relevant features to start with.

Future work on this project will include attempts to automate parts of the process to facilitate the use of larger corpora, and the implementation of the data structures and code necessary to automate the inquiry process.

## ACKNOWLEDGMENTS

This work was done in conjunction with Jim Martin and Susanna Cumming whose help is gratefully acknowledged. It was supported by the National Science Foundation under Contract No. IRI-9109859.